\documentclass[conference]{IEEEtran}
\usepackage{times}

\usepackage[numbers]{natbib}
\usepackage{multicol}

\usepackage{graphicx}
\usepackage{wrapfig}
\usepackage{multirow}
\usepackage{epstopdf}
\usepackage{tabulary}
\usepackage[hidelinks]{hyperref}
\usepackage{mathptmx} 
\usepackage{times} 
\usepackage{amsmath} 
\usepackage{amsfonts}
\usepackage{amssymb}  
\usepackage{color}
\usepackage[bottom]{footmisc} 

\usepackage{color}

\usepackage[T1]{fontenc}
\usepackage{fancyhdr}

\begin{document}

\title{\small \vspace{-1cm} RSS Pioneers 2019 - Freiburg, Germany, June 2019\\ \vspace{0.5cm}\huge Object Perception and Grasping in Open-Ended Domains}



\author{S. Hamidreza Kasaei\\
Department of Artificial Intelligence, University of Groningen, The Netherlands. \\
hamidreza.kasaei@rug.nl -- \href{http://www.ai.rug.nl/hkasaei}{www.ai.rug.nl/hkasaei}}

\maketitle

\IEEEpeerreviewmaketitle

\section{Introduction}
Nowadays service robots are leaving the structured and completely known environments and entering human-centric
settings. For these robots, object perception and grasping are two challenging tasks due to the high demand for
accurate and real-time responses. Although many problems have already been understood and solved successfully, many challenges still remain. \emph{Open-ended learning} is one of these challenges waiting for many improvements. Cognitive science revealed that humans learn to recognize object categories and grasp affordances ceaselessly over time. This ability allows adapting to new environments by enhancing their knowledge from the accumulation of experiences and the conceptualization of new object categories. Inspired by this, an autonomous robot must have the ability to process visual information and conduct learning and recognition tasks in an open-ended fashion. In this context, ``\emph{open-ended}'' implies that the set of object categories to be learned is not known in advance, and the training instances are extracted from online experiences of a robot, and become gradually available over time, rather than being completely available at the beginning of the learning process. 

As an example, consider a cutting task, if the robot does not know what a ``\emph{Knife}'' is, it may ask a user to show one instance and demonstrate how to grasp the ``\emph{Knife}'' to execute the task. Such situations provide opportunities to collect training instances from actual experiences of the robot and the system can incrementally update its knowledge rather than retraining from scratch when a new task is introduced or a new category is added. This way, it is expected that the competence of the robot increases over time. 

In my research, I mainly focus on interactive open-ended learning approaches to recognize multiple objects and their grasp affordances concurrently. In particular, I try to address the following research questions:

\begin{itemize}
\item What is the importance of open-ended learning for autonomous robots?
\item How robots could learn incrementally from their own experiences as well as from interaction with
humans?
\item What are the limitations of Deep Learning approaches to be used in an open-ended manner?
\item How to evaluate open-ended learning approaches and what are the right metrics to do so?
\end{itemize}
 
 \begin{figure}[t]
\center
  \includegraphics[width=\linewidth, trim= 0.cm 0.cm 0.cm 0.3cm,clip=true]{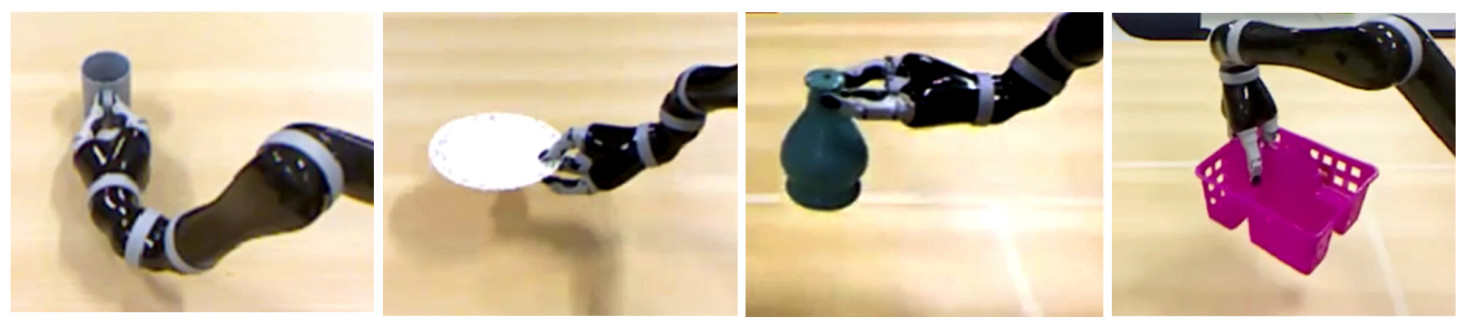}
\vspace{-6mm}
\caption{Four examples of affordance detection results.}
\label{fig:grasp}       
\vspace{-6mm}
\end{figure}

\section {Related Work}
\vspace{-1mm}
Several projects have been conducted to develop robots to assist people in daily tasks. Some examples of service robot platforms that have demonstrated perception and action coupling include Rosie~\cite{beetz2011robotic}, HERB~\cite{srinivasa2010herb}, ARMAR-III~\cite{vahrenkamp2010integrated} and Walk-Man~\cite{tsagarakis2017walk}. Unlike our approach, these projects broadly employ static perception systems to perform object recognition and manipulation tasks. In these cases, the knowledge of robots is fixed, in the sense that the representation of the known object categories or grasp templates does not change after the training stage. Our works enable a robot to improve their knowledge from accumulated experiences through interaction with the environment~\cite{kasaei2015adaptive}, and in particular with humans. A human user may guide the process of experience acquisition, teaching new concepts, or correcting insufficient or erroneous concepts through interaction. 

In recent studies on object recognition and grasp affordance detection, much attention has been given to deep Convolutional Neural Networks (CNNs). It is now clear that if in a scenario,\emph{ we have a fixed set of object categories} and \emph{a massive number of examples per category that are sufficiently similar to the test images}, CNN-based approaches yield good results, notable recent works include \cite{redmon2016you}\cite{levine2016learning}\cite{bousmalis2018using}\cite{mahler2017dex}. In open-ended scenarios, these assumptions are not satisfied, and the robot needs to learn new concepts on-site using very few training examples. While CNNs are very powerful and useful tools, there are several limitations to apply
them in open-ended domains. In general, CNN approaches are incremental by nature but not open-ended, since the
inclusion of new categories enforces a restructuring in the topology of the network. Furthermore, training a CNN-based
approach requires long training times and training with a few examples per category poses a challenge for these methods. Deep transfer learning can relax these limitations and motivate us to combine deeply learned features with an online classifier to handle the problem of open-ended object category learning and recognition.

\section{Approaches and Contributions}
My works contribute in several important ways to the research area of object perception and grasping. The primary goal of my researches is to concurrently learn and recognize objects as well as their associated affordances in an open-ended manner. I organize my works into three main categories and briefly explain the details of each category in the following subsections. All contributions have been evaluated using different standard object and scene datasets and empirically tested on different robotic platforms including PR2 and JACO robotic arm. 

\subsection{Open-Ended Object Category Learning and Recognition}
\begin{figure}[h]
	\vspace{-3mm}
	\centering
	\includegraphics[width=0.9\columnwidth, trim= 1cm 0cm 0cm 0cm,clip=true]{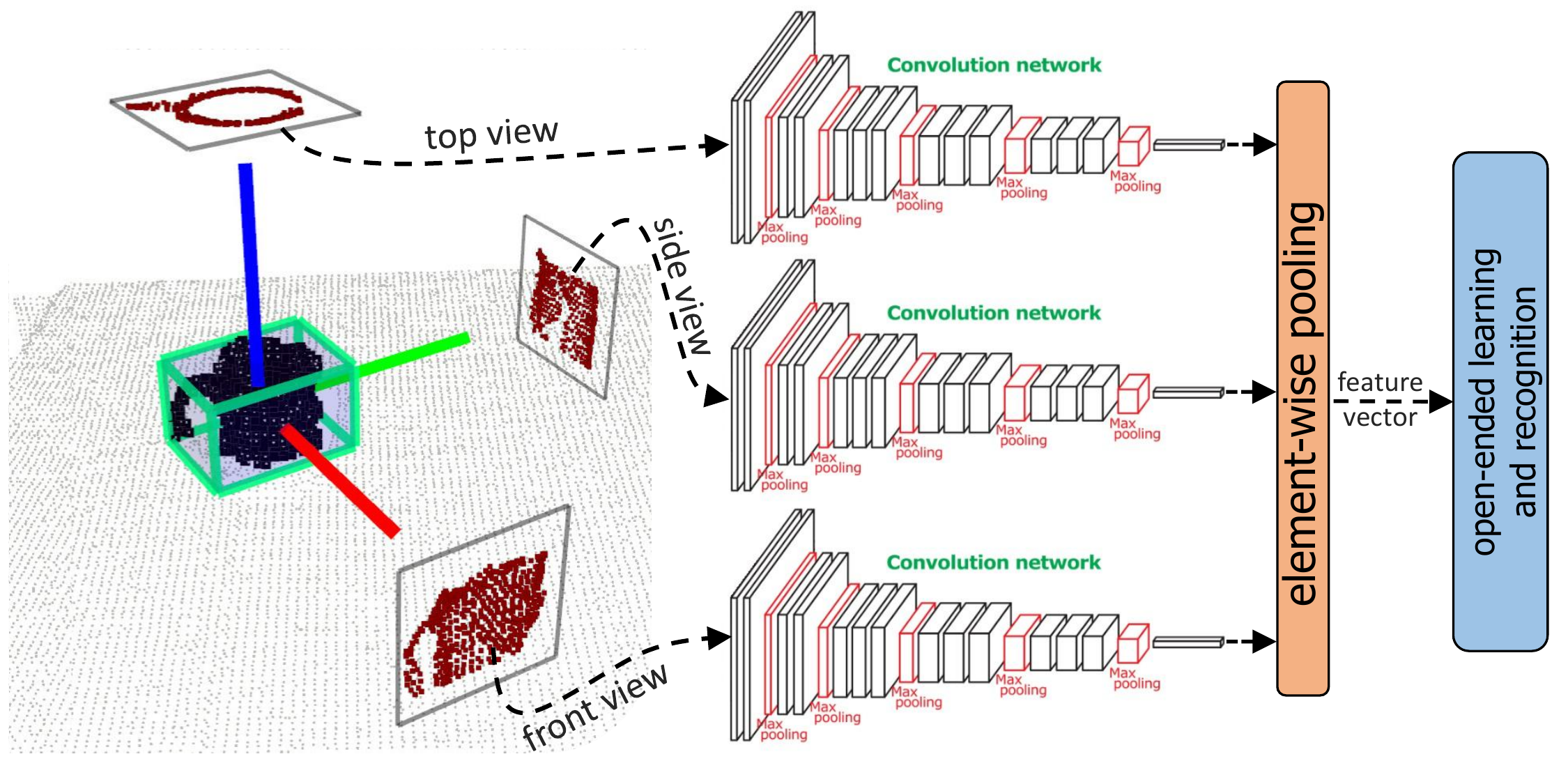}
	\vspace{-4mm}
	\caption{Overview of the proposed \emph{OrthographicNet}.}
	\label{fig:overall}
\end{figure}

Two important parts of my works are concerned with the object representation~\cite{kasaei2016good,kasaei2016hierarchical,oliveira2015concurrent,Sock_2017_ICCV,kasaei2016orthographic,lim2015hierarchical} and object category learning~\cite{kasaei2018towards,oliveira20163d,kasaei2015interactive,oliveira2014perceptual,lim2014interactive}.
In our recent work, we presented a deep transfer learning based approach for 3D object recognition in open-ended domains named \emph{OrthographicNet}. In particular, \emph{OrthographicNet} generates a rotation and scale invariant global feature for a given object, enabling to recognize the same or similar objects seen from different perspectives. As depicted in Fig.~\ref{fig:overall}, we first construct a unique reference frame for the given object. Afterward, three principal orthographic projections including front, top, and right-side views are computed by exploiting the object reference frame. Each projected view is then fed to a CNN, pre-trained on ImageNet~\cite{deng2009imagenet}, to obtain a view-wise deep feature. The obtained view-wise features are then merged, using an element-wise max-pooling function to construct a global feature for the given object. The obtain global feature is scale and pose invariant, informative and stable, and deigned with the objective of supporting accurate 3D object recognition. We finally conducted our experiments with an instance-based learning and a nearest neighbor classification rule. This approach was extensively evaluated in \cite{hamidrezakasaei2019orthographicnet}. Experimental results show that \emph{OrthographicNet} yields significant improvements over the previous state-of-the-art approaches concerning scalability, memory usage and object recognition performance. 

\vspace{-1mm}
\subsection{Grasp Affordance Learning and Recognition}

Robots are still not able to grasp all unforeseen objects and finding a proper grasp configuration, i.e. the position and orientation of the arm relative to the object, is still challenging~\cite{kasaei2016object,shafii2016learning,kasaei2018perceiving}. One approach for grasping unforeseen objects is to recognize an appropriate grasp configuration from previous grasp demonstrations. The underlying assumption in this approach is that new objects that are similar to known ones (i.e. they are familiar) can be grasped in a similar way. Towards this goal, two interconnected learning and recognition approaches are developed. First, an instance-based learning approach is developed to recognize familiar object views; and second, a grasp learning approach is proposed to associate grasp configurations (i.e., end-effector positions and orientations) to grasp affordance categories. The grasp pose learning approach uses local and global visual features of a demonstrated grasp to learn a grasp template associated with the recognized affordance category. It is worth mentioning the grasp affordance category and the grasp configuration are taught through verbal and kinesthetic teaching, respectively. This approach was extensively evaluated in \cite{kasaei2019grasp}. The experimental results demonstrate the high reliability of the developed template matching approach in recognizing the grasp poses. Experimental results also show how the robot can incrementally improve its performance in grasping familiar objects. Fig.~\ref{fig:grasp} shows four examples of our approach.

\subsection{Open-Ended Evaluation}

The number of new open-ended learning algorithms proposed every year is growing rapidly. Although all the proposed methods have been shown to make progress over the previous one, it is challenging to quantify this progress without a concerted evaluation protocol. Moreover, the ability of different open-ended learning techniques to cope with context change in the absence of explicit cueing is not evaluated so far. It is worth mentioning off-line evaluation methodologies such as cross-validation are not well suited to evaluate open-ended learning systems, since they do not comply with the simultaneous nature of learning and recognition in autonomous robots. Moreover, they assume that the set of categories is predefined. We therefore propose a novel experimental evaluation methodology, that takes into account the open-ended nature of object category learning in multi-context scenarios~\cite{kasaei2018coping}. This is an important contribution as published results are usually not comparable. A set of systematic experiments was carried out in~\cite{kasaei2018coping} to thoroughly evaluate and compare a set of state-of-the-art methods in depth, not only in the classic single context setting but also in the multi-context open-ended settings. When an experiment is carried out, learning performance is evaluated using several measures including: (\emph{i}) the number of learned categories at the end of the experiment, an indicator of \emph{How much the system was capable of learning}; (\emph{ii}) the number of question / correction iterations required to learn those categories and the average number of stored instances per category, indicators of \emph{How fast does it learn?} and \emph{How much memory does it take?} respectively; (\emph{iii}) Global Classification Accuracy (GCA), computed using all predictions in a complete experiment, and the Average Protocol Accuracy (APA), i.e., average of all accuracy values successively computed to control the application of the teaching protocol. GCA and APA are indicators of \emph{How well does it learn?}

\section {Future Work}

In the continuation of these works, I would like to investigate the possibility of using deep transfer learning
methods for recognizing 3D object category and grasp affordance concurrently in open-ended domains. Moreover, it
would be interesting to extend the proposed learning methods to other domains such as task-informed grasping and manipulation by addressing \emph{how people can learn new tasks extremely quickly?}

\small{
\bibliographystyle{plainnat}
\bibliography{bib/library,bib/refs_Hamid}
}
\end{document}